\documentclass[preprint,12pt]{elsarticle}

\usepackage{amssymb}

\usepackage{graphicx}
\usepackage{multirow}
\usepackage{float}
\usepackage{amsmath}
\usepackage{amssymb,bm}
\usepackage{subcaption,xcolor}
\usepackage{color, soul}
\usepackage{ctable}
\usepackage{pifont}

\usepackage[colorlinks,linkcolor=black]{hyperref}

\journal{Journal of Systems Architecture}

\begin{document}

\begin{frontmatter}



\title{Fast Monocular Hand Pose Estimation on Embedded Systems}


\author[1]{Shan An}
\author[2]{Xiajie Zhang}
\author[2]{Dong Wei}
\author[1]{Haogang Zhu\corref{mycorrespondingauthor}}
\cortext[mycorrespondingauthor]{Corresponding author}
\author[3]{Jianyu Yang}
\author[4]{Konstantinos A. Tsintotas}
\address[1]{School of Computer Science and Engineering, Beihang University, Beijing, 100191, China.}
\address[2]{Tech. and Data Center, JD.COM Inc, Beijing, 100108, China.}
\address[3]{School of Rail Transportation, Soochow University, Suzhou, 215006, China.}
\address[4]{Department of Production and Management Engineering, Democritus University of Thrace, Xanthi, 67132, Greece.}


\begin{abstract}
Hand pose estimation is a fundamental task in many human-robot interaction-related applications. However, previous approaches suffer from unsatisfying hand landmark predictions in real-world scenes and high computation burden. This paper proposes a fast and accurate framework for hand pose estimation, dubbed as ``FastHand". Using a lightweight encoder-decoder network architecture, FastHand fulfills the requirements of practical applications running on embedded devices. The encoder consists of deep layers with a small number of parameters, while the decoder makes use of spatial location information to obtain more accurate results. The evaluation took place on two publicly available datasets demonstrating the improved performance of the proposed pipeline compared to other state-of-the-art approaches. FastHand offers high accuracy scores while reaching a speed of 25 frames per second on an NVIDIA Jetson TX2 graphics processing unit.
\end{abstract}

%


\end{frontmatter}


\section{Introduction}
\label{sec:introduction}

Gesture recognition is an essential component needed for a robust Human-Robot Interaction (HRI) \cite{brethes2004face, cicirelli2015kinect, ehlers2016human, mazhar2018towards, chang2019improved, kansizoglou2019active}.
Since robot malfunctions may occur due to the noisy sensor measurements or operator's speech disabilities, an alternative technique has to be adopted to allow the agent to understand the human's intentions and, consequently, react accordingly \cite{goldin1999role}.
However, gesture recognition is a challenging problem in the robotics community as the hand is defined through a small area compared to the human body.
Moreover, it exhibits a high degree of freedom and similarity, i.e., among the finger joints' visual appearance.
The timing required for a robot to interact with its operator constitutes another considerable challenge in HRI since real-time responses are necessary for a reliable application.
Skeleton-based methods \cite{nunez2018convolutional, devineau2018deep, nguyen2019neural, liu2020decoupled}, generally follow two processing steps.
At first, they detect hand landmarks, a task widely known as hand pose estimation, and subsequently, they recognize gestures according to these points.
Some approaches make use of depth cameras to locate the 3D hand landmarks \cite{oikonomidis2011efficient, ren2013robust, tagliasacchi2015robust, ge2018robust, wan2019self}.
Nevertheless, these modules are expensive for small and low-cost platforms.
Therefore, it is urgent to study hand pose estimation using a monocular camera as the primary sensing modality.

Based on the technique they use to tackle the problem, modern methods can be divided into the top-down and bottom-up ones.
The former firstly detect the hand and afterward locate its landmarks.
On the contrary, bottom-up frameworks initially detect the landmarks, and next, they cluster them to form a hand.
Both strategies present advantages and disadvantages.
The top-down methods repeatedly run on each hand bounding box to localize landmarks, resulting in the linearly increased run-time.
The bottom-up approaches detect landmarks from the whole image but have a low recall rate because the complex background causes false positives.

Furthermore, hand pose estimation pipelines are divided into two categories according to their output.
Methods that utilize coordinate regression are placed in the first \cite{zimmermann2019freihand, chen2020nonparametric, santavas2020attention}, whereas algorithms that extract heatmaps belong to the second one \cite{ge20193d, zimmermann2017learning, mueller2018ganerated, cai2018weakly, iqbal2018hand, simon2017hand}.
More specifically, approaches that fall into the first category directly predict the landmark coordinates from images.
However, this is not accurate and has limitations when encountering occlusions and large poses.
Frameworks belonging to the second category infer each hand joint's likelihood heatmap and then generate their coordinates.
These techniques offer high accuracy scores; nevertheless, they present the disadvantage of a slower execution time, making them impracticable for resource-constrained mobile platforms.
This is mainly due to their used network structure since they do not follow a lightweight architecture.

\begin{figure}[]
	\begin{center}
   	\includegraphics[width=1.0\linewidth]{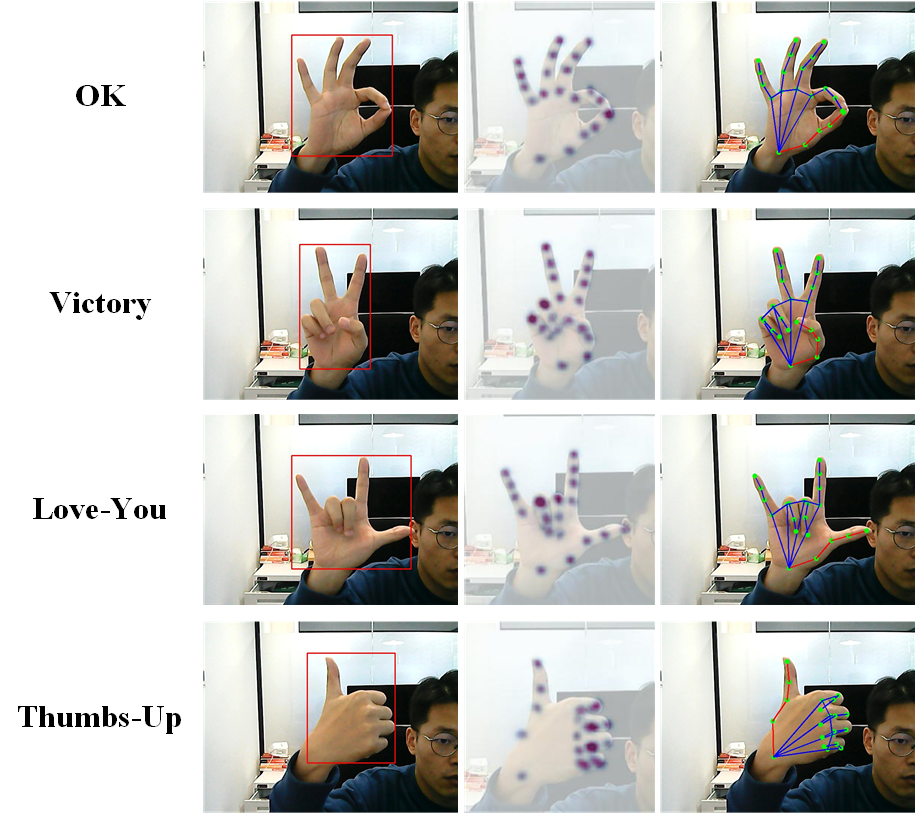} \par
	\end{center}
   \caption{An illustrative example of the proposed hand pose estimation pipeline.
   The first column shows the input frames where the detected hand bounding-boxes are indicated.
   The second column depicts the predicted heatmaps painted in the cropped image, while in the third column, the predicted 2D coordinates for the 21 hand landmarks are demonstrated.}
\label{fig:example}
\end{figure}

Considering the reasons above, we propose a lightweight and top-down pose estimation technique, dubbed ``FastHand" in this paper.
Our method firstly detects the hand using a monocular camera, afterward crops the Region Of Interest (ROI), and then localizes its 2D landmarks based on a heatmap regression (Fig. \ref{fig:example}).
An encoder-decoder network design is employed.
Regarding the encoder, a low amount of parameters are maintained while learning high-level information by using $1\times1$ convolution, skip connection \cite{he2016deep}, and depthwise separable convolutions \cite{chollet2017xception}.
Concerning the decoder, an improvement in feature map resolution is achieved through a deconvolution process, which preserves the high resolution of the hand and avoids the false recognition of the background.
Moreover, the heatmap regression accuracy is also improved.
Extensive experiments are conducted on two publicly available datasets demonstrating our method's superior performance compared to other state-of-the-art frameworks.
The main contributions can be summarized below:
\begin{itemize}
  \item A novel, lightweight encoder-decoder network, entitled ``FastHand", is proposed for hand pose estimation based on heatmap regression.
  The proposed framework is able to perform on a mobile robot in real-time.
  \item A 2D hand landmark dataset is generated based on the Youtube3D Hands image-sequence \cite{kulon2020weakly}. The generation code\footnote{https://github.com/AnshanTJU/Youtube3DHands-2D} is made available to facilitate future studies.
  \item Extensive experiments are performed to demonstrate the effectiveness of the proposed method.
\end{itemize}

The rest of the paper is organized as follows.
Section \ref{relatedwork} reviews the related study about the hand pose estimation task.
In Section \ref{method}, the proposed method is described, while Section \ref{experimental} evaluates the system's performance and reports the comparative results.
Finally, our paper concludes in Section \ref{conclutions}.

\section{RELATED WORK}
\label{relatedwork}

\subsection{Coordinate Regression-Based Approaches}

FreiHAND \cite{zimmermann2019freihand} proposed the first large-scale, multi-view, and 3D annotated real dataset, wherein a Convolutional Neural Network (CNN) was trained \cite{zimmermann2019freihand}.
Subsequently, this framework could predict hand poses from the incoming RGB images with a certain ability of generalization.
NSRM \cite{chen2020nonparametric} adopted a cascaded multi-task architecture aiming to jointly learn the hand structure and its landmarks' representation \cite{chen2020nonparametric}.
A novel CNN with a ``self-attention" module was proposed in \cite{santavas2020attention}.
This network is lightweight and is capable of deployment on embedded systems.
A deep regression network, combined with a specific label distribution learning network, is proposed to learn hand landmarks from 3D point cloud \cite{xu20203d}.
``MediaPipe Hands" \cite{zhang2020mediapipe} first determines the position of hands with a palm detector and then inputs the bounding-box containing hands to the landmark model to regress 21 2.5D hand landmarks directly.

\subsection{Heatmap Regression-Based Approaches}

Methods based on heatmap regression showed an improved performance regarding the final accuracy than the one achieved via the coordinate regression.
In \cite{zimmermann2017learning}, a deep network was proposed to estimate the 3D hand pose originated from RGB images.
This framework detected the hand's location, and then it cropped the selected region.
Subsequently, an encoder-decoder network extracted the 2D score maps for landmarks definition.
Finally, 3D hand poses were estimated from these points by another network trained a priori on 3D data.
A multi-view training method \cite{simon2017hand} used Convolutional Pose Machines (CPM) \cite{wei2016convolutional} as a detector and achieved similar performance to RGB-D systems.
In \cite{wang2018mask}, two stages determine the network's functionality.
Firstly, the hand mask prediction took place, while the pose prediction followed.
For a given RGB image, the hand bounding box and landmarks were obtained simultaneously through a single forward pass of an encoder-decoder network in \cite{wang2019srhandnet}.
Afterward, the hand region was used as feedback to determine whether a cycle detection was performed.
The monocular 3D hand tracking performed well even in challenging occlusion scenarios through the synthetic generation of the training data \cite{mueller2018ganerated}.
Cai \textit{et al.} \cite{cai2018weakly} adopted a weakly-supervised pipeline trained on \mbox{RGB-D} images and predicted 3D joints from the incoming color frames.
Utilizing a CNN for depth and heatmap estimation, 2.5D poses are obtained in \cite{iqbal2018hand}.
Adaptive Graphical Model Network (AGMN) \cite{kong2019adaptive} integrated a CNN with a graphical model for accurate 2D estimation.
Similarly, a Rotation-invariant Mixed Graph Model Network (R-MGMN) performed 2D gesture estimation through a monocular RGB camera \cite{kong2020rotation}.
The authors in \cite{moon2020interhand2} proposed the InterHand2.6M dataset and InterNet for cases where two hands are used.
Boukhayma1 \textit{et al.} introduced the first end-to-end deep learning method, which was formed by an encoder-decoder structure placed in series \cite{boukhayma20193d}.
2D landmark information was utilized to facilitate the learning of a 3D hand pose.
A graph CNN-based method was proposed to estimate 3D hand joints and reconstructed its 3D mesh of the surface in \cite{ge20193d}.
Our network is based on heatmap regression because of its high accuracy.

\section{PROPOSED METHOD}
\label{method}

\begin{figure*}[]
\begin{center}
   \includegraphics[width=0.9\linewidth]{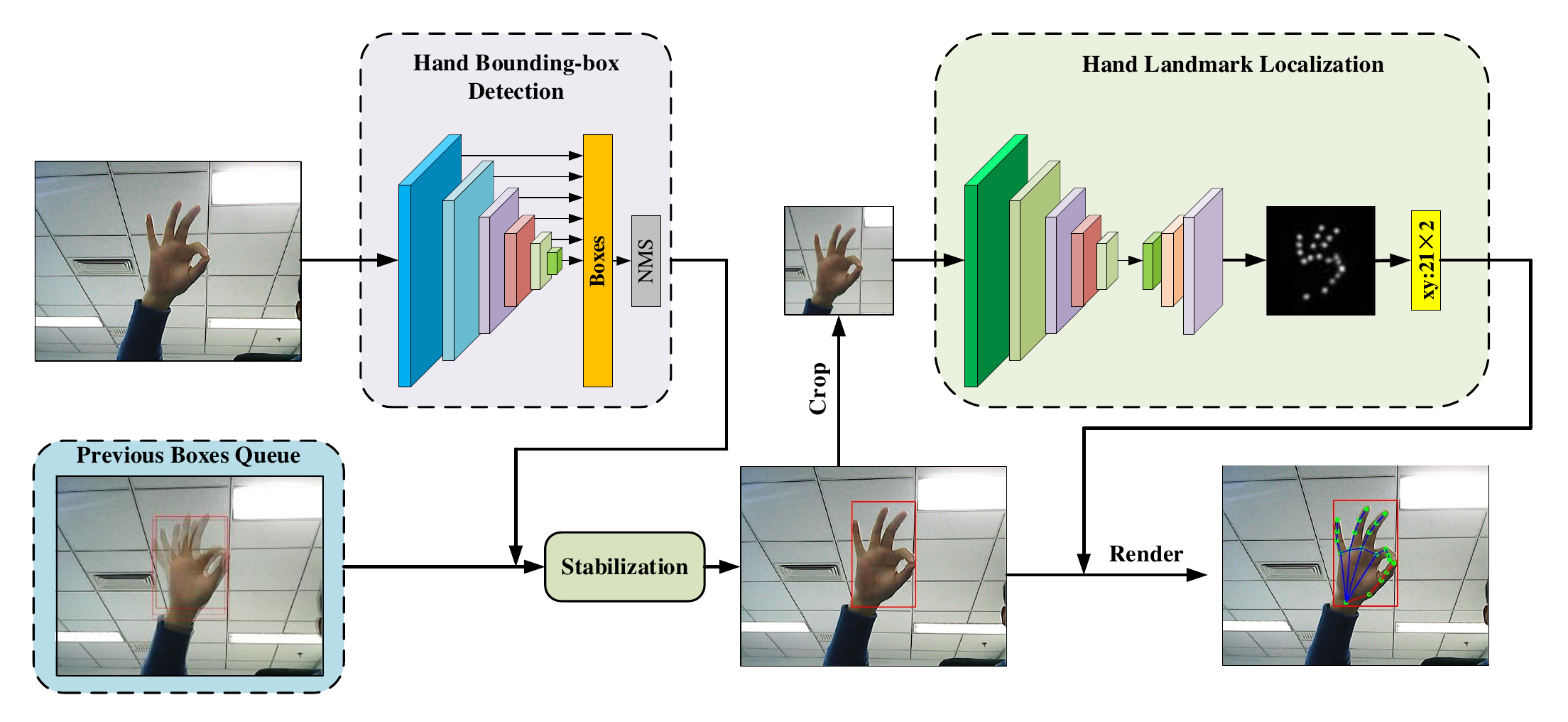} \par
\end{center}
   \caption{An overview of the proposed framework.
Firstly, though the MobileNet-SSD network, which is based on MobileNetV2 \cite{sandler2018mobilenetv2} and Single Shot multi-box Detector (SSD) \cite{liu2016ssd}, we detect the human hand in the incoming RGB image.
Subsequently, aiming to stabilize the Region Of Interest (ROI), this area is combined with the previous visual sensory information.
Next, we crop the ROI and feed it into the landmark localization module.
The encoder-decoder network is used to predict the landmarks' heatmap and compute its 2D position.
	}
\label{fig:system}
\end{figure*}

An overview of the proposed pipeline is outlined in Fig.~\ref{fig:system}.
The input image is fed into the hand detection network to get the corresponding bounding box.
We obtain a stabilized box through the information provided by the previous boxes.
Afterward, this ROI is placed on the landmark localization network for computing the heatmaps.
Finally, $2$D positions of the $21$ hand landmarks are extracted.

\subsection{Hand Detection and Stablization}

\subsubsection{Hand Detection}

Top-down approaches firstly detect the ROI, i.e., the hand, in the visual sensory information, and then localize its landmarks on the chosen area.
In this way, a computational complexity reduction is achieved, which is our main focus.
We select to utilize the MobileNet-SSD framework to detect human hands, which is based on MobileNetV2 \cite{sandler2018mobilenetv2} as the network's backbone and Single Shot multi-box Detector (SSD) \cite{liu2016ssd} for the detection.

\subsubsection{Stabilization of the Bounding-Box}

As we adopt the top-down approach, the detection box's inevitable jitters occur, affecting the landmarks' localization.
Aiming for more accurate prediction scores, we added a fast and straightforward method to stabilize the hand bounding box.
Since early perceived frames present a higher correlation with the current box, we perform a weighted average according to the exponentially decreasing weight.
This way, we exploit the correlation between the current box and the previous ones as the weight's value is higher for neighboring images.
The current bounding-box coordinates $P_{cur}$ are calculated as:
\begin{equation}
P_{cur}=\sum_{k=0}^{n} P_k\times \frac{e^{-k}}{\sum_{j=0}^{n}e^{-j}},
	\label{eq21}
\end{equation}
where $k$ represents the previous frame's index (e.g. $0$ denotes the current frame, while $1$ is the previous one) and $n$ corresponds to the multitude of the examined frames.
This value was set heuristic to 6 ($n$ = 6).
$P_k$ are the coordinates of the $k$ bounding-box.
Finally, the last term of Eqn.~\ref{eq21}, i.e., $\frac{e^{-k}}{\sum_{j=0}^{n}e^{-j}}$, is the exponentially decreasing weight.

\subsection{Landmark Localization}

\begin{figure*}[t!]
\begin{center}
   \includegraphics[width=1.0\linewidth]{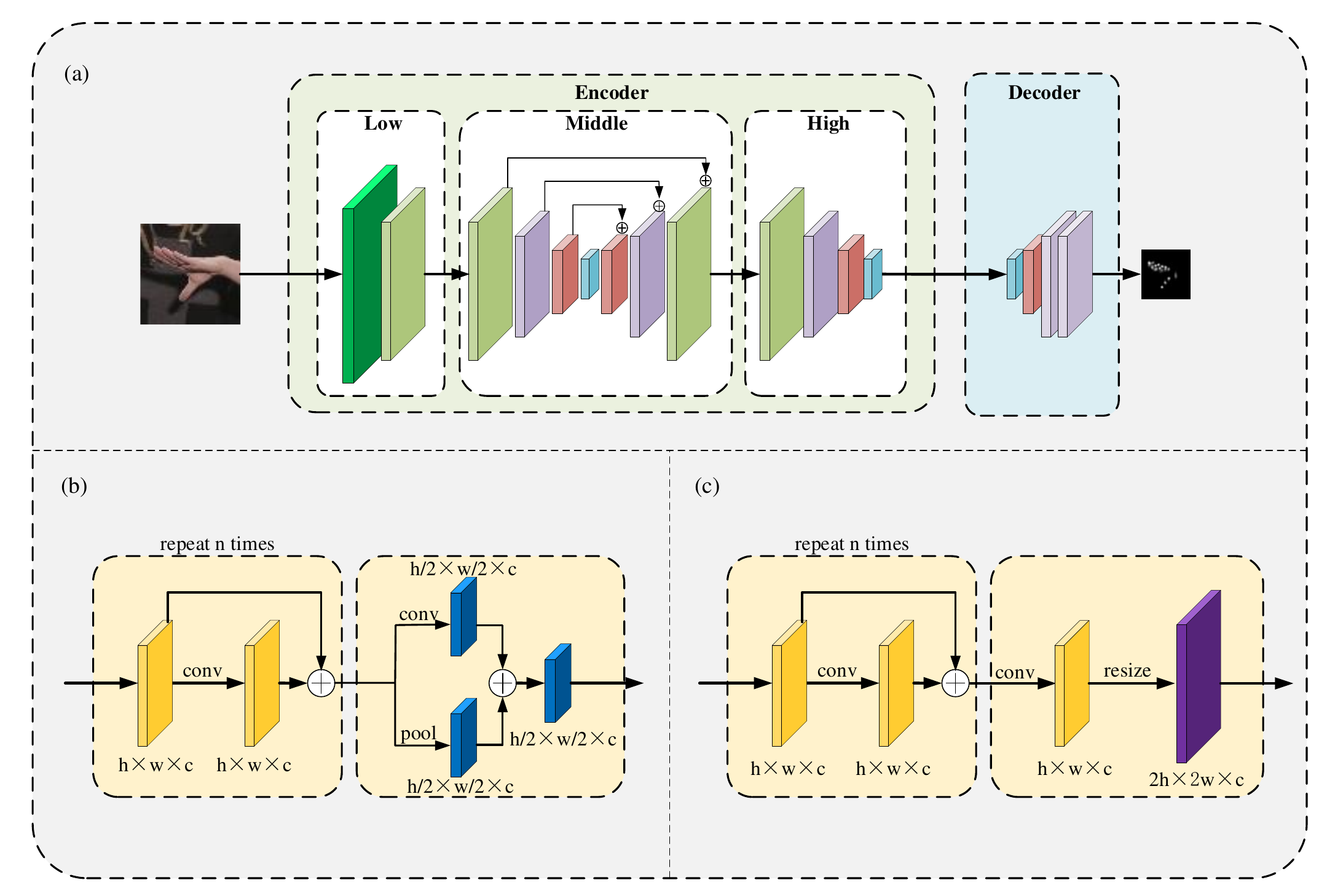} \par
\end{center}
   \caption{The illustration of the proposed landmark localization network.
The encoder-decoder structure is shown in division (a).
The encoder network has three parts (the Low, the Middle, and the High), which encodes the high-level information into low-resolution feature maps using a small number of parameters and computations.
In the decoder, the three deconvolution operations improve the output resolution.
The components in (b) are the downsampling operation between all feature maps with reduced resolution in the encoder.
The convolutions performed in (c) are the upsampling operations of the encoder's middle part to improve the resolution.
   }
\label{fig:network}
\end{figure*}

For the landmarks' detection, we design a heatmap regression method following the encoder-decoder architecture outlined in Fig.~\ref{fig:network}.
At first, resize the image obtained from the previous steps to $256\times256$, and then feed it to the proposed network to extract high-level information.
The feature map resolution is restored to $64\times64$ through the decoder to generate the heatmaps, which are processed to extract the 2D coordinates of the landmarks.

\subsubsection{The Encoder Network}

The encoder network utilizes $1\times1$ convolution, skip connection \cite{he2016deep}, and depthwise separable convolutions \cite{chollet2017xception}, since we aim to a lightweight solution.
The network structure follows the design of  ``MediaPipe Hands" \cite{zhang2020mediapipe}.
The network preserves a small number of parameters while learning more high-level information.
Three parts define its structure, namely the Low, the Middle, and the High (Fig.~\ref{fig:network} (a)).
More specifically, low-level features are extracted via the first part, composed of repeated residual blocks ($\times 4$) and downsampling operation.
In the Middle part, the utilization of parameters is improved through an encoder-decoder network and skip connection.
In the encoder, a downsampling operation is utilized to reduce the resolution with residual blocks repeated four times, as illustrated in Fig.~\ref{fig:network} (b).
Then in the decoder part, the resolution can be improved through the upsampling module, which consists of residual blocks, a convolution, and a resize operation,  as shown in Fig.~\ref{fig:network} (c).
At last, in the High part, more high-level information is extracted while the downsampling module reduces the resolution, as shown in Fig.~\ref{fig:network} (b).

\subsubsection{The Decoder Network}
The decoder network aims to restore the resolution of the feature map.
Since the human hand presents a different structure from the face or other human parts, presenting higher degrees of freedom, we improve the feature map resolution through a deconvolution ($\times 3$) performed to preserve spatial position information.
This technique offers a better modeling capability than a direct, coordinate regression and can provide improved accuracy.
When heatmaps are used as ground truth to predict the landmarks, they must have high resolution.
In our method, the encoder's output is $8\times8$.
As a result, we use the decoder to increase the resolution to $64\times64$.
This way, we manage to retain more position-related information.
Three layers of deconvolutions are used to restore the feature map size.
The proposed pipeline outputs $64\times64\times21$ heatmaps.
Finally, the 2D coordinates of the 21 hand landmarks are derived by extracting the peak points at each channel.

\section{EXPERIMENTS}
\label{experimental}

In this section, extensive experiments are conducted to demonstrate the effectiveness of FastHand.
Due to the lack of the 2D hand pose dataset, we generate a large-scale dataset using image-sequences from YouTube3D Hands \cite{kulon2020weakly}.
Two widely-used datasets, STB \cite{zhang20163d} and RHD \cite{zimmermann2017learning}, are selected for the method's evaluation.
Subsequently, we introduce the experimental settings, which include the training parameters and evaluation metrics.
Finally, we provide quantitative and qualitative results to fully demonstrate the overall performance of our method.

\begin{table*}[]
	\caption{Properties of the used datasets. }
	\label{table_dataset}
	\begin{center}
	\resizebox{\columnwidth}{!}{
	\begin{tabular}{c|c|c|c|c|c}
	\toprule
	\textbf{Dataset} & \textbf{Usage} & \textbf{Description} & \textbf{Image resolution} & \textbf{\# Joints} & \textbf{\# Images} \\
	\midrule
	YouTube2D Hands & Training  & Real-world & $256\times256$ &21 & 47125 \\
	\midrule
    GANeratedHands \cite{mueller2018ganerated} & Training &  Synthetic & $256\times256$ &21 & 141449  \\
    \midrule
    Test set of STB \cite{zhang20163d}& Test & Real-world  & $640\times480$  & 21 & 6000  \\
    \midrule
    Test set of RHD \cite{zimmermann2017learning}& Test & Synthetic  & $320\times320$  & 21 & 2727 \\
	\bottomrule
	\end{tabular}
   }
	\end{center}
\end{table*}

\subsection{2D Hand Pose Dataset Generation}
There are several limitations of the existed 2D hand pose dataset.
The datasets, such as STB \cite{zhang20163d}, MHP \cite{gomez2019large}, HO3D \cite{hampali2019ho}, only have limited scenes and hand movements.
On the other hand, synthetic datasets are always unreal.
It is difficult to collect and annotate a 2D hand pose because it has high degrees of freedom, self-occlusion, and mutual occlusion.

We generate a 2D hand pose dataset based on the YouTube3D Hands \mbox{\cite{kulon2020weakly}} dataset, which is a large-scale image-sequence including hand actions.
It provides 109 videos in total.
102 of them are the training sets, while the rest are the test ones.
As this dataset offers the 3D vertex coordinates of hands, we need to extract the 2D information of the hand landmarks.
Through the average of 10 vertices around each hand skeleton, we get the 3D coordinates of the 21 hand landmarks.
Subsequently, we project them onto the 2D plane to obtain the 2D points.
The generated 2D hand pose dataset is referred to as YouTube2D Hands.
As a final note, the images are resized to $256\times256$ and augmented by 10 times using random scaling and cropping.
Besides, we use the GANeratedHands dataset \cite{mueller2018ganerated} for the network's training since, through the utilization of real-world images, we cannot cover all gestures, and the annotation in real data is always inaccurate.
A brief description of each dataset is shown in Table \ref{table_dataset}.

\subsection{Experimental Settings}

\subsubsection{Training Process}

The training are conducted on four NVIDIA Tesla P40 GPU.
The entire network was trained for a total of 50 epochs, while we decreased the learning rate exponentially after the first three ones.
The initial learning rate was set to $10^{-2}$ and the batch size was set to 64.
The Adam optimizer \cite{kingma2014adam} was utilized as well.
Finally, the network's implementation was based on TensorFlow \cite{abadi2016tensorflow}.

\subsubsection{Evaluation Metrics}

Sum of Squares Error (SSE), \mbox{end-point error} (EPE), and Probability of Correct Keypoint (PCK) within a normalized distance threshold, are used as evaluation metrics:
\begin{equation}
	SSE = \frac{\sum_{s=1}^{D}( {\sum_{i=1}^{21}}((\frac{y_{si}-\widehat{y}_{si}}{max(w,h)} )^2) )}{D},
	\label{eq31}
\end{equation}

\begin{equation}
	EPE = \frac{\sum_{s=1}^{D}( {\sum_{i=1}^{21}}\left \| \frac{(y_{si}-y_{s0})-(\widehat{y}_{si}-\widehat{y}_{s0})}{max(w,h)}  \right \|  )}{21\times D},
	\label{eq32}
\end{equation}

\begin{equation}
	PCK_{\sigma}^i=\frac{1}{D} {\sum_{s=1}^{D}1(\frac{\left \|y_{si}-\widehat{y}_{si} \right \| }{max(w,h)} \le {\sigma})},
	\label{eq33}
\end{equation}

\begin{equation}
	PCK_{\sigma}=\frac{\sum_{i=1}^{21}PCK_{\sigma}^i}{21},
	\label{eq34}
\end{equation}
where the term $y_{si}$ is the ground truth of landmark $i$, and $\widehat{y}_{si}$ is the system's predicted coordinate.
$i$ represents the landmark indexes $i= [1..., 21]$ and $s$ represents the index of hand sample.
$y_{s0}$ and $\widehat{y}_{s0}$ are the ground truth root joint and the predicted root joint.
$ D $ denotes the number of samples in the dataset, while $w$ and $h$ are the original images' width and height, respectively.
In Eq. \ref{eq33}, $1(\cdot)$ is the indicator function and $\sigma$ is the threshold.
If $L2$ distance between the predicted landmark and the ground truth is less than $\sigma$, the indicator function is set to 1, otherwise to 0.
$PCK_{\sigma}^i$ represents the $PCK_{\sigma}$ metric of landmark $i$ at the ratio of $\sigma$, which is an empirical value of $0.2$.

\subsection{Comparison with the State-of-the-art Techniques}

We compare the proposed pipeline against four state-of-the-art methods, namely SRHandNet \cite{wang2019srhandnet}, NSRM Hand \cite{chen2020nonparametric}, MediaPipe Hands \cite{zhang2020mediapipe}, and InterHand \cite{moon2020interhand2}.
As shown in Table~\ref{tab:all}, our FastHand ranks first in most metrics on the STB and RHD datasets.
The proposed approach achieves a $SSE$ score of 0.3490 on the STB dataset, surpassing the second-best (0.4853).
For the $EPE$ metric on the STB dataset, our method is slightly higher than InterHand \cite{moon2020interhand2}.
One can conclude that our method is competitive against state-of-the-art, yielding accurate locating performance of the hand landmarks.

In Tables~\ref{tab:runtime} and \ref{tab:modelsize}, we compare the execution times and the networks' size, respectively.
The execution time reported here is only the time of landmark localization, that is, the model forward time.
As one can observe, our model has a size of 13.0 MB, which is less than 1/3 of SRHandNet.
MediaPipe Hands is the smallest model because the official implementation has many optimizations, such as quantization.
Furthermore, we list the inference runtime for different methods showing that FastHand can execute at 25FPS on an NVIDIA Jetson TX2 GPU and 31FPS on an NVIDIA GeForce 940 MX GPU of a laptop outperforming each other method, without adopting any acceleration.
Note that since the official implementation of MediaPipe Hands is based on TFlite\footnote{https://www.tensorflow.org/lite/guide} with considerable acceleration operations, for a fair evaluation, it is not involved in the inference runtime comparison.

\begin{table*}[]
	\caption{\label{tab:all} {Comparative results of the baseline methods against the proposed method. Green denotes the best, while blue is second best.
$\uparrow$ means the higher the better, while $\downarrow$ means the lower the better. }}	
	\begin{center}
	\resizebox{\columnwidth}{!}{
	\begin{tabular}{c|c|p{2.4cm}<{\centering}|p{2.2cm}<{\centering}|p{2.2cm}<{\centering}|p{2.2cm}<{\centering}|c}
	\toprule
	\textbf{Dataset} & \textbf{Metric} & \textbf{SRHandNet \cite{wang2019srhandnet}} & \textbf{NSRM Hand \cite{chen2020nonparametric}} & \textbf{MediaPipe Hands \cite{zhang2020mediapipe}} & \textbf{InterHand \cite{moon2020interhand2} }  & \textbf{Our Method} \\
	\midrule
	STB \cite{zhang20163d} & $SSE$  $\downarrow$	 &--	 				& 0.7078   &  0.9435  & \textcolor{blue}{0.4853} & {\textcolor{green}{0.3490}}  \\
					 	 & $EPE$ $\downarrow$ & -- 					& 0.1326  &  0.1522  & {\textcolor{green}{0.1302}} & {\textcolor{blue}{0.1317}} \\
 					 	 & $PCK@0.2$ $\uparrow$ & 0.8526 	& 0.7246   &  0.7032 & {\textcolor{blue}{0.8245}} 				   & {\textcolor{green}{0.8948}} \\
	\midrule	
	RHD \cite{zimmermann2017learning} & $SSE$ $\downarrow$ 	 & -- 	 				& 2.5613  &  {\textcolor{blue}{1.9929}} & 2.0413  & {\color{green}{0.6368}} \\
					 	 &  $EPE$ $\downarrow$ & -- 					& {\textcolor{blue}{0.1953}}		 &   	0.2133	& 0.2630 &  {\textcolor{green}{0.0986}} \\
 					 	 & $PCK@0.2$ $\uparrow$ & 0.5317 	& {\textcolor{blue}{0.7177}}   & 0.6927   & {0.3910}   & {\textcolor{green}{0.8661}}	 \\
	\bottomrule
	\end{tabular}
}
	\end{center}
\end{table*}

\begin{table*}[]
	\caption{Comparison in inference runtime (Frames Per Second - FPS) of different pose estimation methods. }
	\label{tab:runtime}
\begin{center}
	\resizebox{\columnwidth}{!}{
	\begin{tabular}{c|c|c|c|c}
	\toprule
	\textbf{Device} & \textbf{SRHandNet \cite{wang2019srhandnet}} & \textbf{NSRM Hand \cite{chen2020nonparametric}} & \textbf{InterHand \cite{moon2020interhand2} } & \textbf{Our Method} \\
	\midrule
    NVIDIA GeForce 940MX GPU & 21.06 FPS  & 2.42 FPS   &   14.24 FPS  & \bfseries{31.33 FPS} \\
	NVIDIA Jetson TX2 GPU & 19.16 FPS & 3.65 FPS & 7.77 FPS & \bfseries{25.05 FPS} \\
	\bottomrule
	\end{tabular}
}
\end{center}
\end{table*}

\begin{table*}[]
	\caption{Comparison in terms of model size of different pose estimation methods. }
	\label{tab:modelsize}
\begin{center}
	\resizebox{\columnwidth}{!}{
	\begin{tabular}{c|c|c|c|c|c}
	\toprule
	\textbf{Model} & \textbf{SRHandNet \cite{wang2019srhandnet}} & \textbf{NSRM Hand \cite{chen2020nonparametric}} & \textbf{MediaPipe Hands \cite{zhang2020mediapipe}} & \textbf{InterHand \cite{moon2020interhand2} } & \textbf{Our Method} \\
	\midrule
	Size (MB) & 71.90 & 139.7 & 3.9 & 541.7 & 13.0 \\
	\bottomrule
	\end{tabular}
}
\end{center}
\end{table*}

\subsection{Qualitative Evaluation}

\begin{figure*}[h!]
	\begin{center}
   	\includegraphics[width=1.0\linewidth]{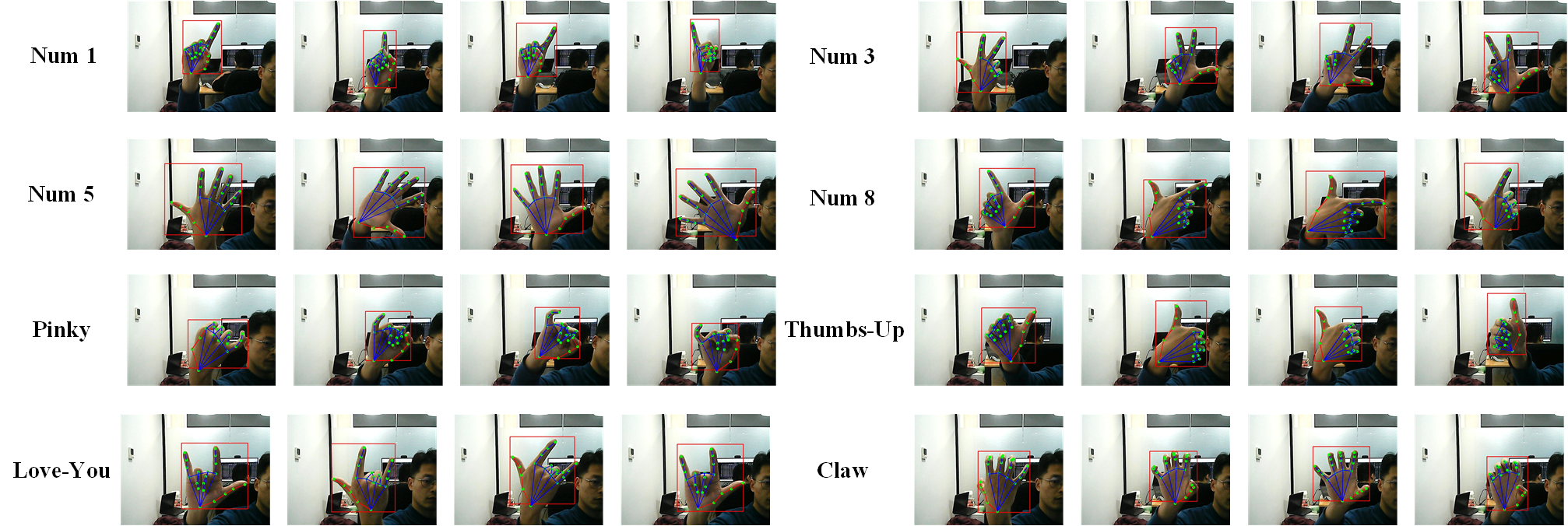} \par
	\end{center}
   	\caption{Visualization results obtained by the proposed FastHand. Different hand poses on real-world scenarios are tested.}
\label{fig:qualitative}
\end{figure*}

\begin{figure}[t!]
	\begin{center}
   	\includegraphics[width=1.0\linewidth]{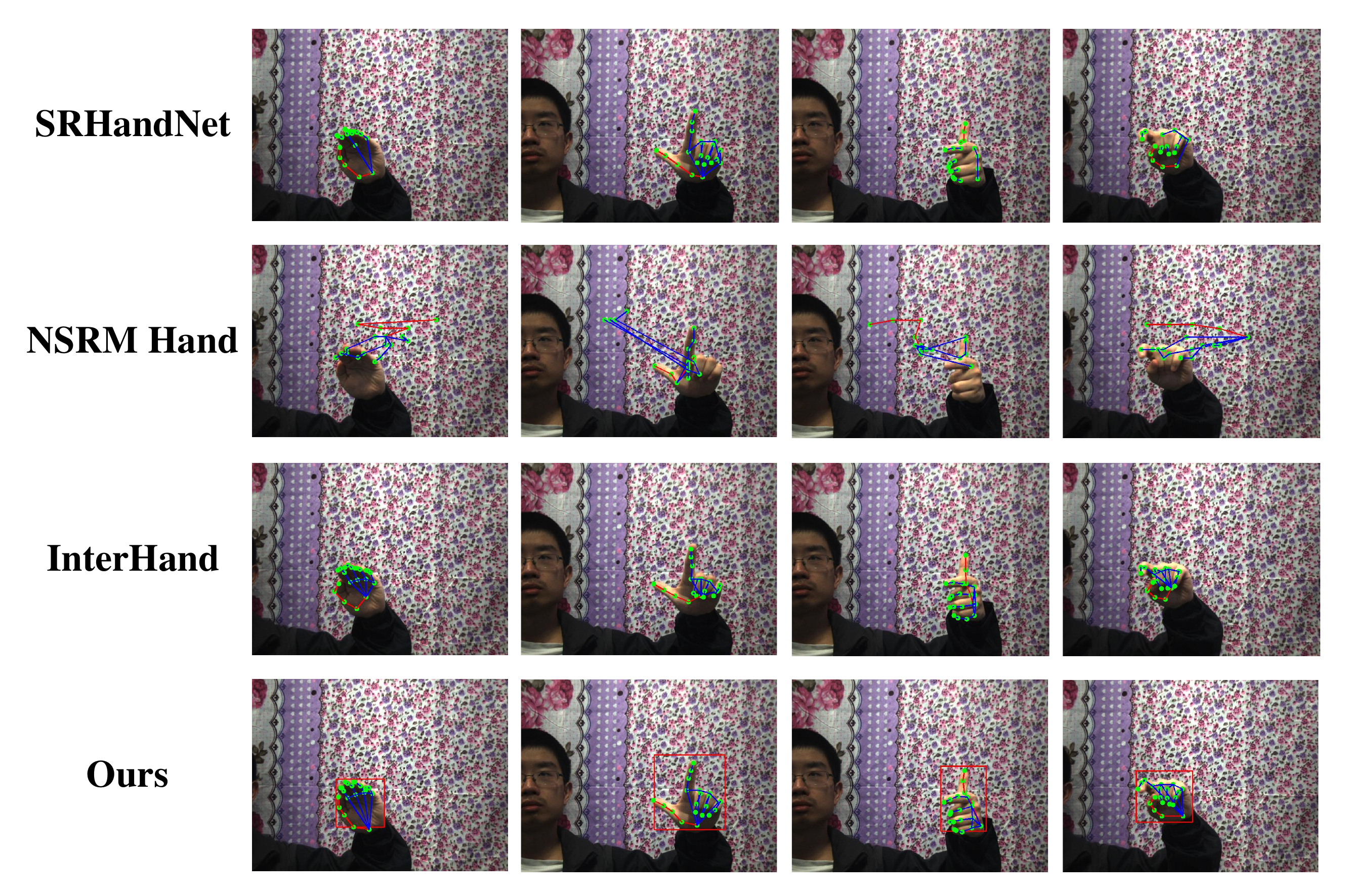} \par
	\end{center}
   	\caption{Qualitative results of the proposed method compared with other approaches on the STB dataset \cite{zhang20163d}.}
\label{fig:qualitative2}
\end{figure}

\begin{figure}[t!]
	\begin{center}
   	\includegraphics[width=1.0\linewidth]{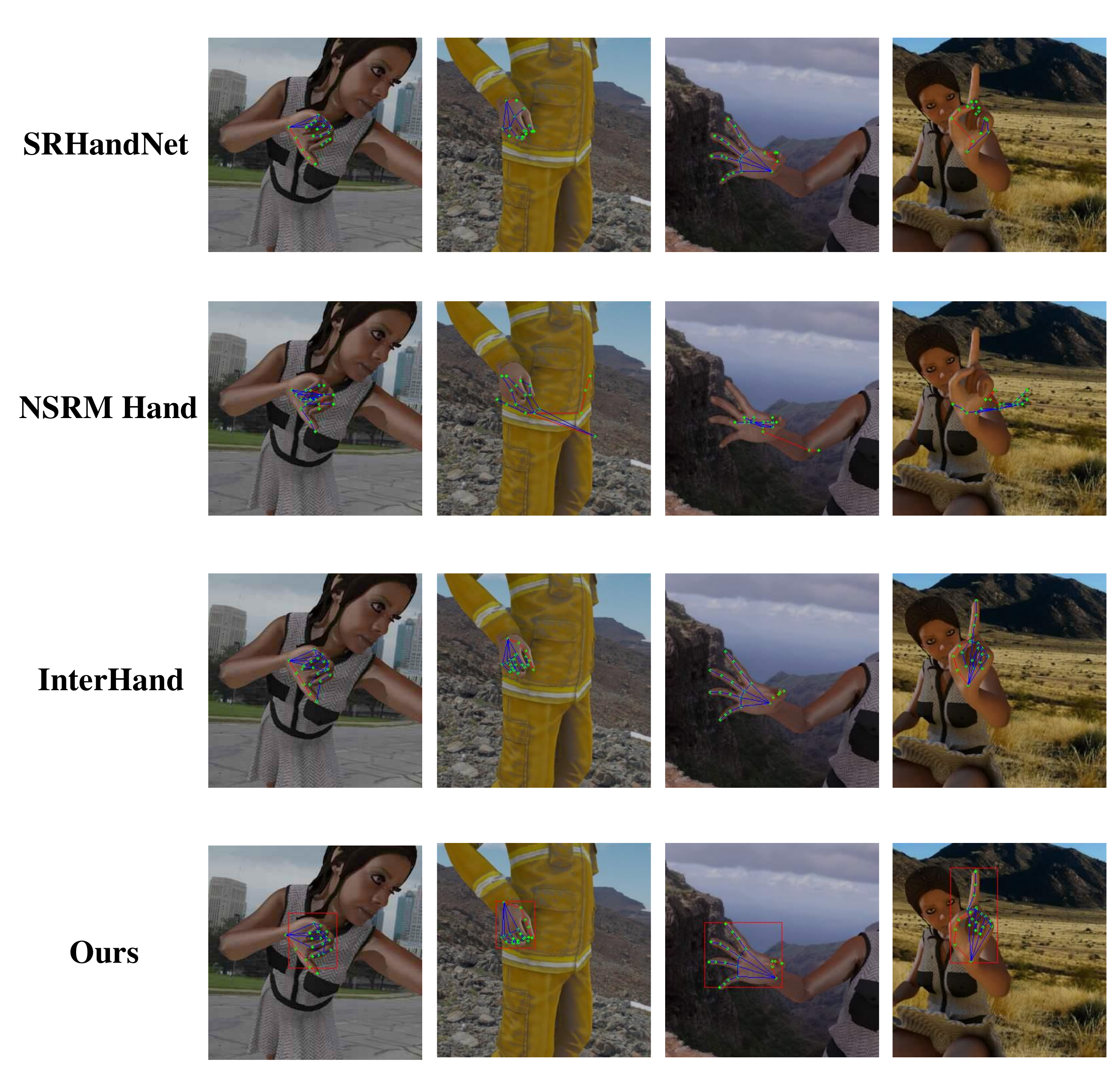} \par
	\end{center}
   	\caption{Qualitative results of the proposed method compared with other approaches on the RHD dataset \cite{zimmermann2017learning}.}
\label{fig:qualitative3}
\end{figure}

We show the visualization results of the proposed pipeline in Fig.~\ref{fig:qualitative}.
Different perspectives of the hand pose are shown in each column.
Our framework can accurately detect the landmarks of open hands, such as ``Num 5", while performs favorably in cases where occlusions occur, such as the fingertips for the pose of ``thumbs-up" and ``claw".
The high-accuracy hand landmark localization is due to our decoder design, which extracts high-resolution feature maps.
In our experiments, we found that our method can recognize the hand gestures of numbers $0$ to $9$ and other basic ones, including the self-occluding poses.
In Fig.~\ref{fig:qualitative2} and Fig.~\ref{fig:qualitative3}, we show the results obtained by FastHand and the other methods on the STB and RHD datasets.
As can be seen, our network predicts accurate coordinates outperforming the rest of the approaches.
The hand pose estimation videos can be found here: \url{https://youtu.be/r9CKDhXXs3Y}.

\section{CONCLUSIONS}
\label{conclutions}

This paper proposes a fast 2D hand pose estimation pipeline, referred to as ``FastHand".
The core of our framework is based on a lightweight encoder-decoder network.
Through hand detection and stabilization, the ROI is cropped and fed into our network for landmark localization.
The proposed network is trained over our large-scale 2D hand pose dataset and a synthetic one while evaluated on two publicly available datasets.
Compared to several state-of-the-art techniques, the proposed approach shows a favorable performance while maintaining a real-time speed (over 25 frames per second) on a low-cost GeForce 940MX GPU and an NVIDIA Jetson TX2 GPU.
Future works can explore the utility of our method in complex robot applications and utilize Part Affinity Fields \cite{cao2017realtime} to differentiate between left hand and right hand.

\section{ACKNOWLEDGEMENT}
This work was supported by grants from the National Key Research and Development Program of China (Grant No. 2020YFC2006200).




\bibliographystyle{elsarticle-num}
\bibliography{ICESS2021}
%
%
%
\end{document}